\documentclass[preprint]{elsarticle}

\usepackage{amssymb}
\usepackage{amsmath}
\usepackage{amsthm}
\usepackage{enumerate}
\usepackage{float}
\usepackage{graphicx}
\usepackage{amsfonts}
\usepackage{mathtools,xcolor,url}
\usepackage{todonotes}
\journal{...}

\newtheorem{theorem}{Theorem}[section]

\newtheorem{lemma}{Lemma}[section]

\newtheorem{cor}{Corollary}[section]
\newtheorem{prop}{Proposition}[section]
\newtheorem{remark}{Remark}[section]

\newtheorem{exam}{Example}[section]

\numberwithin{equation}{section}

\begin{document}

	\begin{frontmatter}

		\title{Neural Network Operators, Shape Properties, Iteration and Associated Markov Chains}

		\author[1]{Ana-Maria Acu} 
		\author[2]{Ioan Ra\c sa}

		\vspace{10cm}
		
		% \author[]{\hspace{10.5cm} }

		\address[1]{Lucian Blaga University of Sibiu, Department of Mathematics and Informatics,  Romania, e-mail: anamaria.acu@ulbsibiu.ro}
			\address[2]{Technical University of Cluj-Napoca, Faculty of Automation and Computer Science, Department of Mathematics, Romania
			e-mail:  ioan.rasa@math.utcluj.ro }

		\begin{abstract} 			{
				
		We construct families of sigmoidal functions and compactly
		supported kernels by recursive averaging and study the
		neural network operators generated by these kernels.
		The construction provides increasing smoothness while
		preserving symmetry, monotonicity, and the partition-of-unity
		property of the kernels. We derive formulas expressing
		the derivatives of the operators in terms of finite
		differences of the sampled function, yielding preservation
		of monotonicity and higher-order convexity on suitable
		interior intervals. We also investigate the iterates of
		both the original operators and their normalized versions.
		For the latter, we determine the limit operator and the
		unique invariant probability measure explicitly.
		The associated finite Markov chains are irreducible,
		aperiodic, and reversible, and their stationary
		distributions give a probabilistic interpretation of
		the limiting operator. For a fixed kernel, these
		stationary distributions approach the uniform
		distribution on the sampling nodes as their number
		tends to infinity.
				 } 
		\end{abstract}

		
	\end{frontmatter}

\section{Introduction}

Approximation by finite linear combinations of affine transforms of a
fixed sigmoidal function provides a classical setting for studying the
approximation capabilities of feedforward neural networks. Cybenko
\cite{a7} showed that, for any continuous sigmoidal function, these
linear combinations are dense in the space of continuous functions on
the unit cube in the uniform norm. Gao and Xu \cite{a8} obtained
quantitative approximation results for univariate functions of bounded
variation. Cardaliaguet and Euvrard \cite{a6} gave an explicit
construction for approximating a function and its derivative.
Subsequent work developed sampling operators based on sigmoidal
activations. Chen and Cao \cite{a2} obtained uniform error estimates for
operators based on the logistic sigmoid, and Costarelli and Spigler
\cite{C-S} studied positive linear neural network operators with
sigmoidal activations on compact intervals.

Several aspects of these operators and their variants have since been
investigated. Voronovskaja-type formulas and high-order convergence were
studied in \cite{COVI-2020}, and Cantarini and Costarelli \cite{a4}
established simultaneous approximation results for a function and its
derivatives. Qian and Yu \cite{QianYu2022} studied rates of approximation
for neural network interpolation operators. For nonlinear variants, Costarelli and Vinti \cite{a3}
studied saturation and inverse results for max-product operators, while
Aslan, De Marchi and Erb \cite{a5} established \(L^p\)-convergence for
Kantorovich-type max-min operators.

The sampling structure of these neural network operators connects them
with the theory of generalized sampling operators. Bardaro and
Mantellini \cite{b3} established a Voronovskaya-type formula for
a general class of discrete operators and applied it to generalized
sampling-type operators. They obtained a quantitative version for
generalized sampling operators in terms of moduli of smoothness
\cite{b2}. They also derived asymptotic formulas for linear
combinations of generalized sampling operators that achieve higher
orders of approximation, with examples including kernels built from
translates of central B-splines \cite{b1}.

The starting point for our shape-preservation results is the work of
Acu, Costarelli, Ra\c sa and Vinti \cite{a1}. They studied approximation
and endpoint behavior for neural network operators without normalization
and obtained shape-preservation results for operators based on the ramp
function and on central B-splines. In particular, their derivative
formula for the spline operators expresses each derivative under
consideration as a weighted sum of finite differences of the sampled
function. We extend this formula to a family of compactly supported
kernels generated by successive averages over intervals of length one.
Each averaging step increases the guaranteed degree of smoothness by one
and preserves symmetry, monotonicity on either side of the origin, and
the partition of unity. We construct a corresponding family of sigmoidal
functions and show that the kernels are their centered differences.
The resulting derivative formula yields preservation of monotonicity
and higher-order convexity on suitable interior subintervals. For a
fixed kernel, the operators also converge to the sampled function
uniformly on compact subintervals of \((-1,1)\) as the grid is refined.

We then fix the sampling grid and study what happens when the
operator is applied repeatedly. Iterates of positive linear and
Markov operators have been studied in \cite{N,R}. In our setting,
truncation of the kernel sum at the endpoints leads to a loss of mass
in the unnormalized sampling matrix. The iterates of the corresponding
operators converge uniformly to zero. Dividing by the sum of the
kernel weights gives operators that preserve constants. When the grid parameter is at least the kernel support radius and the
kernel gives positive weight to adjacent nodes, the normalized iterates converge
uniformly to a constant determined by the unique invariant probability
measure. We give this measure explicitly: its mass at each sampling
node is proportional to the sum of the unnormalized kernel weights at
that node.

The normalized sampling matrix is the transition matrix of a finite
irreducible, aperiodic and reversible Markov chain. Its stationary
distribution consists of the weights of the invariant measure and
thus determines the limit of the iterates. For a fixed kernel, only a
bounded number of rows near the endpoints have an unnormalized row
sum below one. Consequently, the total variation distance between the
stationary distribution and the uniform distribution on the sampling
grid is \(O(n^{-1})\) as \(n\to\infty\).

\section{A family of kernels}\label{s2}

Let \(\varphi_0\in C(\mathbb R)\) be an even function. Suppose that
there exists \(b_0>\dfrac{1}{2}\) such that \(\varphi_0(x)=0\) for
\(|x|\geq b_0\), and that \(\varphi_0\) is increasing on
\([-b_0,0]\) and decreasing on \([0,b_0]\). Assume further that
\[
\sum_{j\in\mathbb Z}\varphi_0(x-j)=1,
\qquad x\in\mathbb R,
\]
and
\[
\int_{\mathbb R}\varphi_0(t)\,dt=1.
\]

For \(k\in\mathbb N_0\), define recursively
\[
\varphi_{k+1}(x)
:=\int_{x-\frac12}^{x+\frac12}\varphi_k(t)\,dt,
\qquad x\in\mathbb R,
\]
and set
\[
b_k:=b_0+\frac{k}{2}.
\]

\begin{prop}
	For every \(k\in\mathbb N_0\), the following properties hold:
	\begin{enumerate}
		\renewcommand{\labelenumi}{(\roman{enumi})}
		\item \(\varphi_k\in C^k(\mathbb R)\);
		\item \(\varphi_k\) is even;
		\item \(\varphi_k(x)=0\) for \(|x|\geq b_k\);
		\item \(\varphi_k\) is increasing on \([-b_k,0]\) and
		decreasing on \([0,b_k]\);
		\item \(\displaystyle\sum_{j\in\mathbb Z}\varphi_k(x-j)=1\)
		for every \(x\in\mathbb R\);
		\item \(\displaystyle\int_{\mathbb R}\varphi_k(t)\,dt=1\);
		\item 
		$\operatorname{supp}\varphi_k=[-b_k,b_k]$.
	\end{enumerate}
\end{prop}

\begin{proof}
	We proceed by induction on \(k\). The assertions hold for \(k=0\)
	by the assumptions on \(\varphi_0\). Suppose that they hold for some
	\(k\in\mathbb N_0\).
Since \(\varphi_k\in C^k(\mathbb R)\), the recursive definition gives
\(\varphi_{k+1}\in C^{k+1}(\mathbb R)\). Thus (i) holds for \(k+1\).

Using the evenness of \(\varphi_k\) and the substitution \(t=-u\), we obtain
\[
\begin{aligned}
	\varphi_{k+1}(-x)
	&=\int_{-x-\frac12}^{-x+\frac12}\varphi_k(t)\,dt \\
	&=\int_{x-\frac12}^{x+\frac12}\varphi_k(-u)\,du \\
	&=\int_{x-\frac12}^{x+\frac12}\varphi_k(u)\,du
	=\varphi_{k+1}(x).
\end{aligned}
\]
Hence (ii) holds for \(k+1\).

Next, let \(x\geq b_{k+1}=b_k+\frac12\). Then
\(t\geq b_k\) for every \(t\in[x-\frac12,x+\frac12]\).
By the induction hypothesis, \(\varphi_k(t)=0\) throughout this
interval, and therefore \(\varphi_{k+1}(x)=0\).
The same conclusion holds for \(x\leq-b_{k+1}\). This proves (iii)
for \(k+1\).

To prove (iv), note first that, by (iii) and (iv) for \(k\),
\(\varphi_k\) is nondecreasing on \((-\infty,0]\) and nonincreasing
on \([0,\infty)\). Moreover,
\begin{equation}\label{eq:kernel-derivative}
	\varphi_{k+1}'(x)
	=\varphi_k\!\left(x+\frac12\right)
	-\varphi_k\!\left(x-\frac12\right),
	\qquad x\in\mathbb R.
\end{equation}
If \(x\leq-\frac12\), both arguments on the right-hand side are
nonpositive. Since \(\varphi_k\) is nondecreasing on
\((-\infty,0]\), we have \(\varphi_{k+1}'(x)\geq0\).

If \(-\frac12\leq x\leq0\), then
\[
0\leq x+\frac12\leq\frac12-x.
\]
By the evenness and the monotonicity of \(\varphi_k\) on
\([0,\infty)\),
\[
\varphi_k\!\left(x+\frac12\right)
-\varphi_k\!\left(x-\frac12\right)
=
\varphi_k\!\left(x+\frac12\right)
-\varphi_k\!\left(\frac12-x\right)
\geq0.
\]
Thus \(\varphi_{k+1}'(x)\geq0\) for every \(x\leq0\).
Therefore,
\[
\varphi_{k+1}'(x)\geq 0,
\qquad x\leq 0.
\]

We now show that the inequality is strict for
\(x\in(-b_{k+1},0)\). Since \(b_0>\frac12\) and $b_k=b_0+\frac{k}{2}$, it follows that \(b_k>\frac12\) for every \(k\in\mathbb N_0\).

Let \(x\in(-b_{k+1},0)\). If \(x\leq-\frac12\), then
\[
x-\frac12<x+\frac12\leq0
\quad\text{and}\quad
x+\frac12>-b_k.
\]
Since \(\varphi_k\) vanishes on \((-\infty,-b_k]\) and is
strictly increasing on \([-b_k,0]\), we obtain
\[
\varphi_k\!\left(x-\frac12\right)
<\varphi_k\!\left(x+\frac12\right).
\]

If \(-\frac12<x<0\), then
\[
0<x+\frac12<\frac12-x
\quad\text{and}\quad
x+\frac12<b_k.
\]
By the evenness of \(\varphi_k\) and its strict decrease on
\([0,b_k]\),
\[
\varphi_k\!\left(x-\frac12\right)
=\varphi_k\!\left(\frac12-x\right)
<\varphi_k\!\left(x+\frac12\right).
\]
Hence, by \eqref{eq:kernel-derivative},
\[
\varphi_{k+1}'(x)>0,
\qquad x\in(-b_{k+1},0).
\]
Since \(\varphi_{k+1}\) is even, it is strictly decreasing on
\([0,b_{k+1}]\). This proves (iv) for \(k+1\).

Next, let \(x\in\mathbb R\). Using the change of variable
\(t=u-j\) and the induction hypothesis, we find
\[
\begin{aligned}
	\sum_{j\in\mathbb Z}\varphi_{k+1}(x-j)
	&=\sum_{j\in\mathbb Z}
	\int_{x-j-\frac12}^{x-j+\frac12}\varphi_k(t)\,dt \\
	&=\int_{x-\frac12}^{x+\frac12}
	\sum_{j\in\mathbb Z}\varphi_k(u-j)\,du \\
	&=\int_{x-\frac12}^{x+\frac12}1\,du
	=1.
\end{aligned}
\]
Thus (v) holds for \(k+1\).

Finally, changing the order of integration yields
\[
\begin{aligned}
	\int_{\mathbb R}\varphi_{k+1}(x)\,dx
	&=\int_{\mathbb R}
	\left(
	\int_{x-\frac12}^{x+\frac12}\varphi_k(t)\,dt
	\right)dx \\
	&=\int_{\mathbb R}\varphi_k(t)
	\left(
	\int_{t-\frac12}^{t+\frac12}dx
	\right)dt \\
	&=\int_{\mathbb R}\varphi_k(t)\,dt
	=1.
\end{aligned}
\]
Thus (vi) holds for \(k+1\). By (iii),
\(\varphi_{k+1}\) vanishes outside
\([-b_{k+1},b_{k+1}]\). Since it is 
increasing on \([-b_{k+1},0]\) and 
decreasing on \([0,b_{k+1}]\), it is positive
throughout \((-b_{k+1},b_{k+1})\). Therefore,
\[
\operatorname{supp}\varphi_{k+1}
=[-b_{k+1},b_{k+1}],
\]
which proves (vii) and completes the induction.
\end{proof}

\section{A family of sigmoidal functions}

Let \(\sigma_0\in C(\mathbb R)\) satisfy the following conditions.
There exists \(a_0>0\) such that
\[
\sigma_0(x)=0 \quad \text{for } x\leq-a_0,
\qquad
\sigma_0(x)=1 \quad \text{for } x\geq a_0.
\]
Moreover, \(\sigma_0\) is increasing on
\([-a_0,a_0]\), the function \(\sigma_0-\frac12\) is odd,
and \(\sigma_0\) is convex on \((-\infty,0]\) and concave on
\([0,\infty)\).

For \(k\in\mathbb N_0\), define recursively
\[
\sigma_{k+1}(x)
:=\int_{x-\frac12}^{x+\frac12}\sigma_k(t)\,dt,
\qquad x\in\mathbb R,
\]
and set
\[
a_k:=a_0+\frac{k}{2}.
\]

\begin{prop}\label{p3.1}
	For every \(k\in\mathbb N_0\), the following properties hold:
	\begin{enumerate}
		\renewcommand{\labelenumi}{(\roman{enumi})}
		\item \(\sigma_k\in C^k(\mathbb R)\);
		\item \(\sigma_k(x)=0\) for \(x\leq-a_k\), and
		\(\sigma_k(x)=1\) for \(x\geq a_k\);
		\item \(\sigma_k\) is  increasing on
		\([-a_k,a_k]\);
		\item \(\sigma_k-\frac12\) is odd;
		\item \(\sigma_k\) is convex on \((-\infty,0]\) and
		concave on \([0,\infty)\).
	\end{enumerate}
	
\end{prop}

\begin{proof}
	We proceed by induction on \(k\). For \(k=0\), properties
	(i)--(v) follow from the assumptions on \(\sigma_0\).
	Suppose that they hold for some \(k\in\mathbb N_0\).
Since \(\sigma_k\in C^k(\mathbb R)\), the recursive definition
implies that \(\sigma_{k+1}\in C^{k+1}(\mathbb R)\). Thus (i)
holds for \(k+1\).

Let \(x\geq a_{k+1}=a_k+\frac12\). Then \(t\geq a_k\)
for every \(t\in[x-\frac12,x+\frac12]\), so
\(\sigma_k(t)=1\) throughout this interval. Hence
\[
\sigma_{k+1}(x)
=\int_{x-\frac12}^{x+\frac12}\sigma_k(t)\,dt
=1.
\]
Similarly, \(\sigma_{k+1}(x)=0\) when \(x\leq-a_{k+1}\).
This proves (ii) for \(k+1\).

Next, \(\sigma_k\) is nondecreasing on \(\mathbb R\), and
\[
\sigma_{k+1}'(x)
=\sigma_k\!\left(x+\frac12\right)
-\sigma_k\!\left(x-\frac12\right)
\geq0,
\qquad x\in\mathbb R.
\]
We claim that this inequality is strict whenever
\(x\in(-a_{k+1},a_{k+1})\). Indeed, if
\(\sigma_{k+1}'(x)=0\), then
\[
\sigma_k\!\left(x+\frac12\right)
=\sigma_k\!\left(x-\frac12\right).
\]
Since \(\sigma_k\) is  increasing on
\([-a_k,a_k]\) and constant outside this interval,
equality is possible only if
\[
x+\frac12\leq-a_k
\quad\text{or}\quad
x-\frac12\geq a_k.
\]
These inequalities imply, respectively,
\(x\leq-a_{k+1}\) or \(x\geq a_{k+1}\), a contradiction.
Therefore,
\[
\sigma_{k+1}'(x)>0,
\qquad x\in(-a_{k+1},a_{k+1}),
\]
and (iii) holds for \(k+1\).

Finally, the induction hypothesis that
\(\sigma_k-\frac12\) is odd gives
\[
\sigma_k(-t)+\sigma_k(t)=1,
\qquad t\in\mathbb R.
\]
By the substitution \(t=-u\), we obtain
\[
\begin{aligned}
	\sigma_{k+1}(-x)+\sigma_{k+1}(x)
	&=\int_{-x-\frac12}^{-x+\frac12}\sigma_k(t)\,dt
	+\int_{x-\frac12}^{x+\frac12}\sigma_k(t)\,dt \\
	&=\int_{x-\frac12}^{x+\frac12}
	\bigl(\sigma_k(-u)+\sigma_k(u)\bigr)\,du \\
	&=\int_{x-\frac12}^{x+\frac12}1\,du
	=1.
\end{aligned}
\]
Thus \(\sigma_{k+1}-\frac12\) is odd, proving (iv)
for \(k+1\).
It remains to prove (v). Since \(\sigma_k\) is continuous,
\(\sigma_{k+1}\in C^1(\mathbb R)\), and
\[
\sigma_{k+1}'(x)
=\sigma_k\!\left(x+\frac12\right)
-\sigma_k\!\left(x-\frac12\right).
\]
We show that \(\sigma_{k+1}'\) is nondecreasing on
\((-\infty,0]\).

If \(x<y\leq-\frac12\), the convexity of \(\sigma_k\)
on \((-\infty,0]\) yields
\[
\sigma_k\!\left(y+\frac12\right)
-\sigma_k\!\left(x+\frac12\right)
\geq
\sigma_k\!\left(y-\frac12\right)
-\sigma_k\!\left(x-\frac12\right).
\]
Thus \(\sigma_{k+1}'(x)\leq\sigma_{k+1}'(y)\).

If \(-\frac12\leq x<y\leq0\), then, using (iv) and
the concavity of \(\sigma_k\) on \([0,\infty)\), we obtain
\[
\begin{aligned}
	\sigma_{k+1}'(y)-\sigma_{k+1}'(x)
	&=
	\sigma_k\!\left(\frac12+y\right)
	-\sigma_k\!\left(\frac12+x\right)\\
	&\quad-
	\sigma_k\!\left(\frac12-x\right)
	+\sigma_k\!\left(\frac12-y\right)
	\geq0.
\end{aligned}
\]

If \(x<-\frac12<y\leq0\), the two cases above give
\[
\sigma_{k+1}'(x)
\leq \sigma_{k+1}'\!\left(-\frac12\right)
\leq \sigma_{k+1}'(y).
\]
Consequently, \(\sigma_{k+1}'\) is nondecreasing on
\((-\infty,0]\), and hence \(\sigma_{k+1}\) is convex there.
By (iv), \(\sigma_{k+1}'\) is even, so it is
nonincreasing on \([0,\infty)\). Therefore,
\(\sigma_{k+1}\) is concave on \([0,\infty)\).
This proves (v) for \(k+1\) and completes the induction.

Consequently, \(\sigma_{k+1}'\) is nondecreasing on
\((-\infty,0]\), so \(\sigma_{k+1}\) is convex there.
By (iv), \(\sigma_{k+1}'\) is even and hence
nonincreasing on \([0,\infty)\). Therefore
\(\sigma_{k+1}\) is concave on \([0,\infty)\).
This proves (v) for \(k+1\) and completes the induction.
\end{proof}

\begin{remark}
	Concerning property (ii) in Proposition \ref{p3.1} see \cite{a1}.
\end{remark}

We now give examples of functions \(\sigma_0\) satisfying
the assumptions of this section.

\begin{exam}The linear ramp provides a simple example:
\[
\sigma_0(x)=
\begin{cases}
	0, & x\leq-\dfrac12,\\[2mm]
	x+\dfrac12, & -\dfrac12<x<\dfrac12,\\[2mm]
	1, & x\geq\dfrac12.
\end{cases}
\]
Here \(a_0=\frac12\).\end{exam}

\begin{exam} For \(\beta\geq\frac12\), let
\(b_\beta=\arcsin\bigl(\frac{1}{2\beta}\bigr)\).
The function
\[
\sigma_0(x)=
\begin{cases}
	0, & x\leq-b_\beta,\\[2mm]
	\dfrac12+\beta\sin x,
	& -b_\beta<x<b_\beta,\\[2mm]
	1, & x\geq b_\beta
\end{cases}
\]
also satisfies the required assumptions.
\end{exam}

\begin{exam}\label{e3.3}Let \(\sigma:\mathbb R\to(0,1)\) be an  increasing
sigmoidal function such that
\(\sigma(-x)=1-\sigma(x)\). Suppose also that
\(\sigma\) is convex on \((-\infty,0]\) and concave
on \([0,\infty)\). For \(\mu>1\), put
\[
c_\mu
=\sigma^{-1}\!\left(\frac{\mu+1}{2\mu}\right)
\]
and define
\[
\sigma_0(x)=
\begin{cases}
	0, & x\leq-c_\mu,\\[2mm]
	\dfrac12+\mu\left(\sigma(x)-\dfrac12\right),
	& -c_\mu<x<c_\mu,\\[2mm]
	1, & x\geq c_\mu.
\end{cases}
\]
Then \(\sigma_0\) satisfies the assumptions of this
section, with \(a_0=c_\mu\).
\end{exam}
The next two examples are particular cases of Example \ref{e3.3}

\begin{exam}For \(\alpha>1\), let
\(a_\alpha=\tan\bigl(\frac{\pi}{2\alpha}\bigr)\).
Another admissible sigmoidal function is
\[
\sigma_0(x)=
\begin{cases}
	0, & x\leq-a_\alpha,\\[2mm]
	\dfrac12+\dfrac{\alpha}{\pi}\arctan x,
	& -a_\alpha<x<a_\alpha,\\[2mm]
	1, & x\geq a_\alpha.
\end{cases}
\]
\end{exam}

\begin{exam} For the logistic function
\[
\sigma(x)=\frac{1}{1+e^{-x}},
\]
we have, with $\mu>1$
\[
c_\mu=\log\frac{\mu+1}{\mu-1}.
\]
Thus, on \((-c_\mu,c_\mu)\),
\[
\sigma_0(x)
=\frac12+
\frac{\mu(1-e^{-x})}{2(1+e^{-x})}.
\]
%Figure~\ref{fig:truncated-logistic} shows this function for \(\mu=2\).
\end{exam}

\begin{figure}[H]
	\centering
	\includegraphics[width=\textwidth]{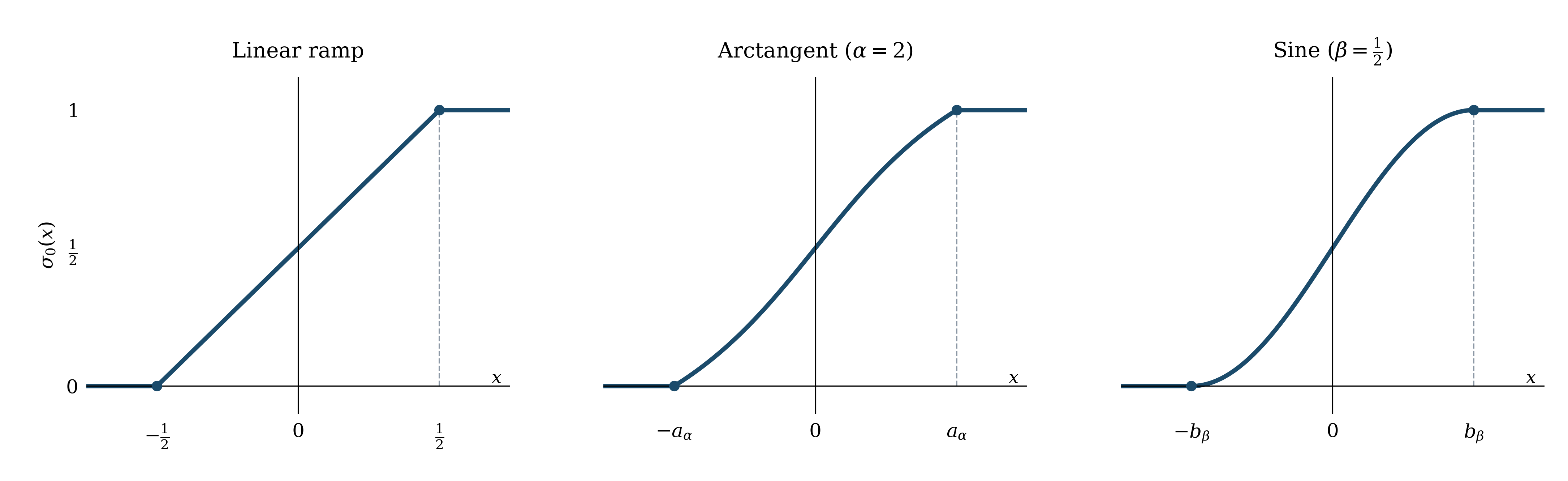}
	\caption{Examples of admissible sigmoidal functions:
		the linear ramp, the arctangent function with \(\alpha=2\),
		and the sine function with \(\beta=\frac12\).}
	\label{fig:sigmoidal-examples}
\end{figure}

\begin{figure}[H]
	\centering
	\includegraphics[width=0.8\textwidth]{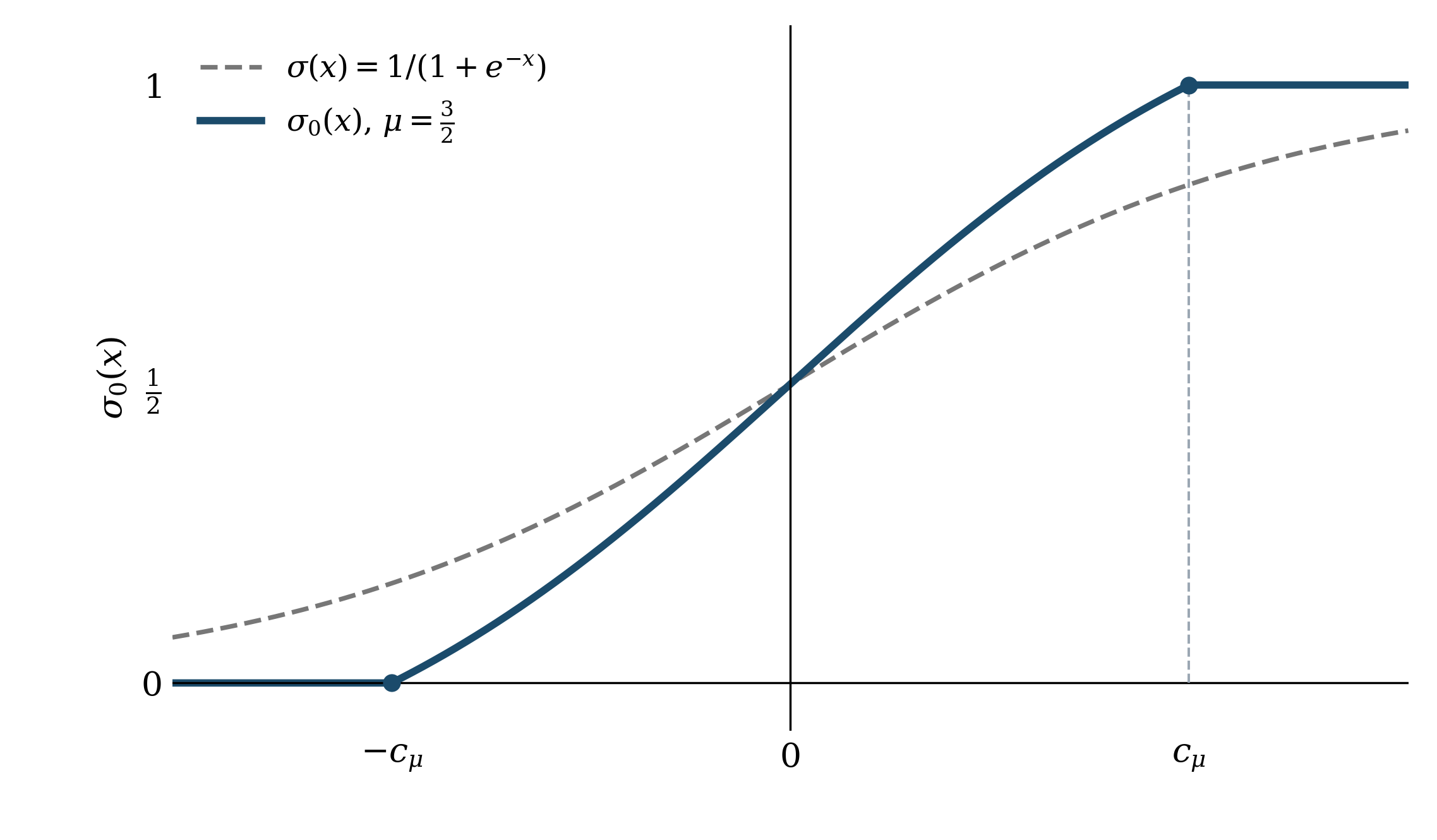}
	\caption{The logistic function and the corresponding
		function \(\sigma_0\) for \(\mu=\frac32\).}
	\label{fig:truncated-logistic}
\end{figure}

Each of these functions denoted by $\sigma_0$ is continuous,  increasing
between its two transition points, and satisfies
\(\sigma_0(-x)=1-\sigma_0(x)\). It is convex on
\((-\infty,0]\) and concave on \([0,\infty)\).
Their graphs are shown in Figure~\ref{fig:sigmoidal-examples} and Figure~\ref{fig:truncated-logistic}.

\begin{prop}
	For every \(k\in\mathbb N\),
	\[
	\sigma_{k-1}(x)\leq\sigma_k(x)
	\quad\text{for }x\leq0,
	\qquad
	\sigma_{k-1}(x)\geq\sigma_k(x)
	\quad\text{for }x\geq0.
	\]
	In particular, \(\sigma_{k-1}(0)=\sigma_k(0)=\frac12\).
\end{prop}

\begin{proof}
	Since \(\sigma_{k-1}-\frac12\) is odd,
	\[
	\sigma_{k-1}(-u)=1-\sigma_{k-1}(u),
	\qquad u\in\mathbb R,
	\]
	and \(\sigma_{k-1}(0)=\frac12\).
	
	Fix \(s,t\geq0\). We claim that
	\begin{equation}\label{eq:symmetric-concavity}
		\sigma_{k-1}(-s+t)+\sigma_{k-1}(-s-t)
		\geq 2\sigma_{k-1}(-s).
	\end{equation}
	If \(t\leq s\), concavity on \([0,\infty)\) gives
	\[
	\sigma_{k-1}(s-t)+\sigma_{k-1}(s+t)
	\leq 2\sigma_{k-1}(s).
	\]
	Applying the symmetry identity to all three terms
	yields \eqref{eq:symmetric-concavity}.
	
	If \(t>s\), concavity implies that the increment
	over \([t-s,t+s]\) is no greater than the increment
	over \([0,2s]\), since these intervals have the same
	length and \(t-s>0\). Hence
	\[
	\begin{aligned}
		\sigma_{k-1}(t+s)-\sigma_{k-1}(t-s)
		&\leq \sigma_{k-1}(2s)-\sigma_{k-1}(0)\\
		&\leq 2\sigma_{k-1}(s)-1.
	\end{aligned}
	\]
	Here the last step follows from concavity at the
	midpoint \(s\). Using
	\(\sigma_{k-1}(-s-t)=1-\sigma_{k-1}(s+t)\) and
	\(2\sigma_{k-1}(-s)=2-2\sigma_{k-1}(s)\),
	we again obtain \eqref{eq:symmetric-concavity}.
	
	By the definition of \(\sigma_k\), splitting the
	integral at \(0\) after the substitution \(u=s+t\)
	gives
	\[
	\begin{aligned}
		\sigma_k(-s)
		&=\int_{-1/2}^{1/2}\sigma_{k-1}(-s+u)\,du\\
		&=\int_0^{1/2}
		\bigl(\sigma_{k-1}(-s+u)
		+\sigma_{k-1}(-s-u)\bigr)\,du\\
		&\geq\int_0^{1/2}2\sigma_{k-1}(-s)\,du\\
		&=\sigma_{k-1}(-s).
	\end{aligned}
	\]
	
	This proves the assertion for \(x\leq0\). For \(x\geq0\),
	applying the result at \(-x\) and using the symmetry
	of both functions gives
	\[
	1-\sigma_{k-1}(x)
	=\sigma_{k-1}(-x)
	\leq\sigma_k(-x)
	=1-\sigma_k(x).
	\]
	Thus \(\sigma_{k-1}(x)\geq\sigma_k(x)\) for \(x\geq0\).
\end{proof}

\section{Sigmoidal functions and kernels}

We retain the notation introduced in the preceding two sections.
The following proposition relates the sigmoidal functions
\(\sigma_k\) to the kernels \(\varphi_k\).

\begin{prop}\label{prop:sigmoidal-kernels}
	Let
	\begin{equation}\label{eq:phi0-sigma0}
	\varphi_0(x)
	=\sigma_0\!\left(x+\frac12\right)
	-\sigma_0\!\left(x-\frac12\right),
	\qquad x\in\mathbb R.
\end{equation}
	Then, for every \(k\in\mathbb N_0\) and \(x\in\mathbb R\),
		\begin{equation}\label{eq.m1}
		\varphi_k(x)
		=\sigma_k\!\left(x+\frac12\right)
		-\sigma_k\!\left(x-\frac12\right);
	\end{equation}
	\begin{equation}\label{eq.m2}
		\sigma_{k+1}(x)
		=\int_{-\infty}^{x}\varphi_k(t)\,dt.
	\end{equation}
\end{prop}

\begin{proof}
	We first prove (\ref{eq.m1}) by induction. It holds for \(k=0\)
	by definition. Suppose that it holds for some
	\(k\in\mathbb N_0\). Then
	\[
	\begin{aligned}
		\varphi_{k+1}(x)
		&=\int_{x-\frac12}^{x+\frac12}
		\varphi_k(t)\,dt\\
		&=\int_{x-\frac12}^{x+\frac12}
		\sigma_k\!\left(t+\frac12\right)\,dt
		-\int_{x-\frac12}^{x+\frac12}
		\sigma_k\!\left(t-\frac12\right)\,dt\\
		&=\int_x^{x+1}\sigma_k(u)\,du
		-\int_{x-1}^{x}\sigma_k(u)\,du\\
		&=\sigma_{k+1}\!\left(x+\frac12\right)
		-\sigma_{k+1}\!\left(x-\frac12\right).
	\end{aligned}
	\]
	Thus (\ref{eq.m1}) holds for every \(k\in\mathbb N_0\).
	
	By (\ref{eq.m1}) and a change of variables,
	\[
	\begin{aligned}
		\int_{-\infty}^{x}\varphi_k(t)\,dt
		&=\int_{-\infty}^{x+\frac12}\sigma_k(u)\,du
		-\int_{-\infty}^{x-\frac12}\sigma_k(u)\,du\\
		&=\int_{x-\frac12}^{x+\frac12}
		\sigma_k(u)\,du
		=\sigma_{k+1}(x).
	\end{aligned}
	\]
	This proves (\ref{eq.m2}).
\end{proof}

 From now on, we assume that \eqref{eq:phi0-sigma0} holds.
Accordingly, the identities \eqref{eq.m1} and \eqref{eq.m2}
hold throughout the remainder of the paper. We also assume that
\(b_0=a_0+\frac12\). Since
\(a_k=a_0+\frac{k}{2}\) and \(b_k=b_0+\frac{k}{2}\), we have
\[
b_k=a_k+\frac12,
\qquad k\in\mathbb N_0.
\]

\section{Neural network operators}

Let \(k,n\in\mathbb N\), and let \(\varphi_k\) be the kernel
introduced in Section \ref{s2}. We consider the operator
\[
\widetilde L_{n,k}:C[-1,1]\longrightarrow C[-1,1],
\]
defined by
\begin{equation}\label{eq:nn-operator}
	(\widetilde L_{n,k}f)(x)
	=\sum_{j=-n}^{n}
	\varphi_k(nx-j)f\!\left(\frac{j}{n}\right),
	\qquad x\in[-1,1].
\end{equation}

Since
\[
\varphi_k(y)
=\int_{y-\frac12}^{y+\frac12}\varphi_{k-1}(t)\,dt,
\]
we have
\[
\varphi_k'(y)
=\varphi_{k-1}\!\left(y+\frac12\right)
-\varphi_{k-1}\!\left(y-\frac12\right).
\]
Consequently, for \(f\in C[-1,1]\),

\begin{align}
	(\widetilde L_{n,k}f)'(x)
	&=n\sum_{j=-n}^{n}
	\varphi_{k-1}\!\left(nx-j+\frac12\right)
	f\!\left(\frac{j}{n}\right)\nonumber\\
	&\quad
	-n\sum_{j=-n}^{n}
	\varphi_{k-1}\!\left(nx-j-\frac12\right)
	f\!\left(\frac{j}{n}\right).\label{eq.r}
\end{align}
Using this formula we prove

\begin{prop}\label{prop:nn-derivative}
	Let \(k,n\in\mathbb N\) and \(f\in C[-1,1]\). Then, for
	\(x\in[-1,1]\),
	\begin{equation}\label{eq:nn-derivative}
		\begin{aligned}
			(\widetilde L_{n,k}f)'(x)
			&=n\varphi_{k-1}\!\left(nx+n+\frac12\right)f(-1)\\
			&\quad
			+n\sum_{j=-n+1}^{n}
			\varphi_{k-1}\!\left(nx-j+\frac12\right)
			\left[
			f\!\left(\frac{j}{n}\right)
			-f\!\left(\frac{j-1}{n}\right)
			\right]\\
			&\quad
			-n\varphi_{k-1}\!\left(nx-n-\frac12\right)f(1).
		\end{aligned}
	\end{equation}
	In particular, if 
	\[
	x\in I_{n,k}:=\left[
	-1+\frac{b_{k}-1}{n},
	\,
	1-\frac{b_{k}-1}{n}
	\right],\,\, n>b_k-1,
	\]
	then
	\begin{equation}\label{eq:nn-derivative-interior}
		(\widetilde L_{n,k}f)'(x)
		=n\sum_{j=-n+1}^{n}
		\varphi_{k-1}\!\left(nx-j+\frac12\right)
		\left[
		f\!\left(\frac{j}{n}\right)
		-f\!\left(\frac{j-1}{n}\right)
		\right].
	\end{equation}
\end{prop}

\begin{proof}
	Formula \eqref{eq:nn-derivative} follows by reindexing the
	second sum in (\ref{eq.r}). For \(x\) in the stated
	interval,
	\[
	nx+n+\frac12\geq b_{k-1},
	\qquad
	nx-n-\frac12\leq-b_{k-1}.
	\]
	Both boundary terms in \eqref{eq:nn-derivative} therefore vanish,
	because \(\varphi_{k-1}(t)=0\) for \(|t|\geq b_{k-1}\).
	This proves \eqref{eq:nn-derivative-interior}.
\end{proof}

We next derive a formula for higher-order derivatives of
\(\widetilde L_{n,k}f\). 

\begin{prop}\label{prop:nn-second-derivative}
	Let \(k\geq2\), \(n\in\mathbb N\), and \(f\in C[-1,1]\).
	For every \(x\in I_{n,k}\),
	\begin{equation}\label{eq:nn-second-derivative}
		\begin{aligned}
			(\widetilde L_{n,k}f)''(x)
			=n^2\sum_{j=-n+2}^{n}
			&\left[
			f\!\left(\frac{j}{n}\right)
			-2f\!\left(\frac{j-1}{n}\right)
			+f\!\left(\frac{j-2}{n}\right)
			\right]\\
			&\qquad{}\times
			\varphi_{k-2}(nx-j+1).
		\end{aligned}
	\end{equation}
\end{prop}

\begin{proof}
	Differentiating \eqref{eq:nn-derivative-interior} gives
	\[
	\begin{aligned}
		(\widetilde L_{n,k}f)''(x)
		=n^2\sum_{j=-n+1}^{n}
		&\left[
		f\!\left(\frac{j}{n}\right)
		-f\!\left(\frac{j-1}{n}\right)
		\right]\\
		&\quad{}\times
		\left[
		\varphi_{k-2}(nx-j+1)
		-\varphi_{k-2}(nx-j)
		\right].
	\end{aligned}
	\]
	After reindexing the second sum, the boundary terms contain
	\(\varphi_{k-2}(nx+n)\) and
	\(\varphi_{k-2}(nx-n)\). They vanish for \(x\in I_{n,k}\).
	Indeed, \(b_k=b_{k-2}+1\) implies
	\[
	nx+n\geq b_{k-2},
	\qquad
	nx-n\leq-b_{k-2}.
	\]
	The remaining terms give \eqref{eq:nn-second-derivative}.
\end{proof}

For \(h>0\), denote by \(\Delta_h^m\) the \(m\)-th forward
difference:
\[
\Delta_h^m f(y)
=\sum_{\ell=0}^{m}
(-1)^{m-\ell}\binom{m}{\ell}f(y+\ell h).
\]

\begin{theorem}\label{thm:nn-higher-derivatives}
	Let \(k,n\in\mathbb N\), \(0\leq m\leq k\), and
	\(f\in C[-1,1]\). Then, for every \(x\in I_{n,k}\),
	\begin{equation}\label{eq:nn-higher-derivatives}
		(\widetilde L_{n,k}f)^{(m)}(x)
		=n^m\sum_{j=-n+m}^{n}
		\varphi_{k-m}\!\left(nx-j+\frac{m}{2}\right)
		\Delta_{1/n}^m f\!\left(\frac{j-m}{n}\right).
	\end{equation}
	In particular, if \(f\in C^m[-1,1]\) and
	\(f^{(m)}\geq0\) on \([-1,1]\), then
	\[
	(\widetilde L_{n,k}f)^{(m)}(x)\geq0,
	\qquad x\in I_{n,k}.
	\]
\end{theorem}

\begin{proof}
	Formula \eqref{eq:nn-higher-derivatives} follows by induction
	on \(m\). The case \(m=0\) is the definition of
	\(\widetilde L_{n,k}\), while the first two steps are given
	by \eqref{eq:nn-derivative-interior} and
	\eqref{eq:nn-second-derivative}. At each subsequent step,
	we use
	\[
	\varphi_r'(y)
	=\varphi_{r-1}\!\left(y+\frac12\right)
	-\varphi_{r-1}\!\left(y-\frac12\right)
	\]
	and reindex one of the resulting sums. The boundary terms
	vanish on \(I_{n,k}\) by the support property of
	\(\varphi_{r-1}\), and the remaining coefficients combine
	into forward differences of the next order.
	
	Finally, if \(f^{(m)}\geq0\), then
	\(\Delta_{1/n}^m f(y)\geq0\) whenever
	\(y,y+m/n\in[-1,1]\). Since
	\(\varphi_{k-m}\geq0\), every summand in
	\eqref{eq:nn-higher-derivatives} is nonnegative.
\end{proof}
 
 \begin{remark}
 The recursive construction of the kernels \(\varphi_k\)
 allows us to obtain operators that preserve higher-order
 shape properties. More precisely, for \(m\leq k\), the
 nonnegativity of \(f^{(m)}\) implies the nonnegativity of
 \((\widetilde L_{n,k}f)^{(m)}\) on \(I_{n,k}\). The trade-off is
 that, for fixed \(n\), this interval becomes smaller as
 \(k\) increases, since \(b_k=b_0+k/2\).
\end{remark} 

We now study the images of the first monomials under
\(\widetilde L_{n,k}\). As usual, let \(e_m(x)=x^m\).

\begin{prop}\label{prop:nn-constant}
	Let \(k,n\in\mathbb N\). For every \(x\in[-1,1]\),
	\begin{equation}\label{eq:nn-constant}
		(\widetilde L_{n,k}e_0)(x)
		=
		\sigma_k\!\left(nx+n+\frac12\right)
		-
		\sigma_k\!\left(nx-n-\frac12\right).
	\end{equation}
	Consequently, if \(n>b_k\), then
	\[
	(\widetilde L_{n,k}e_0)(x)=1,
	\qquad
	x\in I_{n,k}.
	\]
\end{prop}

\begin{proof}
	Using \eqref{eq.m1}, we obtain a telescoping sum:
	\[
	\begin{aligned}
		(\widetilde L_{n,k}e_0)(x)
		&=\sum_{j=-n}^{n}\varphi_k(nx-j)\\
		&=\sum_{j=-n}^{n}
		\left[
		\sigma_k\!\left(nx-j+\frac12\right)
		-\sigma_k\!\left(nx-j-\frac12\right)
		\right]\\
		&=
		\sigma_k\!\left(nx+n+\frac12\right)
		-\sigma_k\!\left(nx-n-\frac12\right).
	\end{aligned}
	\]
	For \(x\in I_{n,k}\), the relation
	\(b_k=a_k+\frac12\) gives
	\[
	nx+n+\frac12\geq a_k,
	\qquad
	nx-n-\frac12\leq-a_k.
	\]
	The two values of \(\sigma_k\) in
	\eqref{eq:nn-constant} are therefore \(1\) and \(0\),
	respectively.
\end{proof}

We denote by \(\Pi_{m-1}\) the space of polynomials
of degree at most \(m-1\).

\begin{prop}\label{prop:nn-monomial}
	Let \(k,n\in\mathbb N\), \(n>b_k\), and
	\(0\leq m\leq k\). There exists
	\(p_{n,k,m}\in\Pi_{m-1}\) such that
	\begin{equation}\label{eq:nn-monomial}
		(\widetilde L_{n,k}e_m)(x)
		=x^m+p_{n,k,m}(x),
		\qquad x\in J_{n,k}:=\left[-1+\frac{b_k}{n},1-\frac{b_k}{n}\right],
	\end{equation}
	where \(p_{n,k,0}=0\).
\end{prop}

\begin{proof}
	The case \(m=0\) follows from
	Proposition~\ref{prop:nn-constant}. Let
	\(1\leq m\leq k\). Since
	\[
	\Delta_{1/n}^{m}e_m(y)=\frac{m!}{n^m},
	\]
	formula \eqref{eq:nn-higher-derivatives} yields
	\[
	(\widetilde L_{n,k}e_m)^{(m)}(x)
	=
	m!\sum_{j=-n+m}^{n}
	\varphi_{k-m}\!\left(nx-j+\frac m2\right).
	\]
	
	We claim that the sum on the right-hand side equals
	\(1\) for \(x\in J_{n,k}\). Indeed, using
	\(b_k=b_{k-m}+\frac m2\), we have
	\[
	nx+n-\frac m2\geq b_{k-m},
	\qquad
	nx-n+\frac m2\leq-b_{k-m}.
	\]
	Thus every nonzero term in
	\[
	\sum_{j\in\mathbb Z}
	\varphi_{k-m}\!\left(nx-j+\frac m2\right)
	\]
	occurs for \(-n+m\leq j\leq n\).
	By the partition of unity property,
	\[
	\sum_{j=-n+m}^{n}
	\varphi_{k-m}\!\left(nx-j+\frac m2\right)=1.
	\]
	It follows that
	\[
	(\widetilde L_{n,k}e_m)^{(m)}(x)=m!,
	\qquad x\in J_{n,k},
	\]
	and this proves \eqref{eq:nn-monomial}.
\end{proof}

\begin{cor}\label{cor:nn-linear}
	Let \(k,n\in\mathbb N\) and \(n>b_k\). Then
	\[
	(\widetilde L_{n,k}e_1)(x)=x,
	\qquad x\in J_{n,k}.
	\]
\end{cor}

\begin{proof}
	By Proposition~\ref{prop:nn-monomial},
	\((\widetilde L_{n,k}e_1)(x)=x+c\) on \(J_{n,k}\)
	for some constant \(c\). Since \(0\in J_{n,k}\)
	and \(\varphi_k\) is even,
	\[
	c=(\widetilde L_{n,k}e_1)(0)
	=\sum_{j=-n}^{n}\frac{j}{n}\varphi_k(j)=0.
	\]
\end{proof}

\begin{prop}\label{prop:nn-evenness}
	If \(f\in C[-1,1]\) is even, then
	\(\widetilde L_{n,k}f\) is even.
\end{prop}

\begin{proof}
	Changing the index \(j\) to \(-j\) and using
	the evenness of \(f\) and \(\varphi_k\), we find
	\[
	\begin{aligned}
		(\widetilde L_{n,k}f)(-x)
		&=\sum_{j=-n}^{n}
		f\!\left(\frac{j}{n}\right)\varphi_k(-nx-j)\\
		&=\sum_{j=-n}^{n}
		f\!\left(-\frac{j}{n}\right)\varphi_k(nx-j)\\
		&=(\widetilde L_{n,k}f)(x).
	\end{aligned}
	\]
\end{proof}

Define the second discrete moment of the kernel by
\[
Q_k:=\sum_{j\in\mathbb Z}j^2\varphi_k(j)=\sum_{j=-b_k}^{b_k}j^2\varphi_k(j).
\]

\begin{cor}\label{cor:nn-quadratic}
	Let \(k,n\in\mathbb N\) and \(n>b_k\). Then
	\[
	(\widetilde L_{n,k}e_2)(x)
	=x^2+\frac{Q_k}{n^2},
	\qquad x\in J_{n,k}.
	\]
\end{cor}

\begin{proof}
	Proposition~\ref{prop:nn-monomial} gives
	\[
	(\widetilde L_{n,k}e_2)(x)=x^2+cx+d,
	\qquad x\in J_{n,k}.
	\]
	By Proposition~\ref{prop:nn-evenness}, the left-hand
	side is even, so \(c=0\). Evaluating at \(x=0\),
	we obtain
	\[
	d=(\widetilde L_{n,k}e_2)(0)
	=\frac1{n^2}\sum_{j=-b_k}^{b_k}j^2\varphi_k(j)
	=\frac{Q_k}{n^2}.
	\]
	\end{proof}

\begin{theorem}\label{thm:nn-convergence}
	For every fixed \(k\in\mathbb N\) and every
	\(f\in C[-1,1]\),
	\[
	\lim_{n\to\infty}(\widetilde L_{n,k}f)(x)=f(x),
	\qquad x\in(-1,1).
	\]
	The convergence is uniform on every compact
	subinterval of \((-1,1)\).
\end{theorem}

\begin{proof}
	Let \(K\) be a compact subinterval of \((-1,1)\).
	For all sufficiently large \(n\), we have
	\(K\subset J_{n,k}\). On this interval, the positive
	linear operators \(\widetilde L_{n,k}\) satisfy
	\[
	\widetilde L_{n,k}e_0=e_0,\qquad
	\widetilde L_{n,k}e_1=e_1,\qquad
	\widetilde L_{n,k}e_2=e_2+\frac{Q_k}{n^2}.
	\]
	Thus their images of the three Korovkin test functions
	converge uniformly on \(K\) to \(e_0,e_1,e_2\),
	respectively. The local form of Korovkin's theorem
	gives uniform convergence to \(f\) on \(K\).
\end{proof}

\section{Iterates of the neural network operators}
\label{sec:nn-iterates}

Fix \(k,n\in\mathbb N\). In this section, we write
\(\widetilde L_n=\widetilde L_{n,k}\) and
\(\varphi=\varphi_k\). Thus,
\[
(\widetilde L_n f)(x)
=
\sum_{j=-n}^{n}
f\!\left(\frac{j}{n}\right)\varphi(nx-j),
\qquad x\in[-1,1].
\]
Set \(I_n=\{-n,-n+1,\ldots,n\}\) and define
\[
u_{n,j}(x):=\varphi(nx-j),
\qquad j\in I_n,\quad x\in[-1,1].
\]
Then
\[
(\widetilde L_n u_{n,j})(x)
=
\sum_{i\in I_n}\varphi(i-j)u_{n,i}(x).
\]

Consider the matrix
\[
A_n:=\bigl(\varphi(i-j)\bigr)_{i,j\in I_n}
\]
and the vectors
\[
U_n(x):=\bigl(u_{n,j}(x)\bigr)_{j\in I_n},
\qquad
F_n(f):=\left(f\!\left(\frac{j}{n}\right)\right)_{j\in I_n},
\]
where \(U_n(x)\) is a row vector and \(F_n(f)\) is a column
vector. The preceding identity gives, by induction,
\begin{equation}\label{eq:nn-iterate-matrix}
	(\widetilde L_n^{\,\ell}f)(x)
	=
	U_n(x)A_n^{\ell-1}F_n(f),
	\qquad \ell\in\mathbb N.
\end{equation}

The matrix \(A_n\) is symmetric because \(\varphi\) is even.
It is also substochastic: its entries are nonnegative and,
by the partition-of-unity property of \(\varphi\),
\[
\sum_{j\in I_n}\varphi(i-j)
\leq
\sum_{j\in\mathbb Z}\varphi(i-j)
=1,
\qquad i\in I_n.
\]
In fact, its spectral radius is strictly less than one.

\begin{prop}\label{prop:nn-matrix-spectrum}
	Every eigenvalue of \(A_n\) belongs to \((-1,1)\).
	Consequently,
	\[
	\lim_{\ell\to\infty}A_n^\ell=0.
	\]
\end{prop}

\begin{proof}
	Since \(A_n\) is symmetric and substochastic, every eigenvalue
	\(\lambda\) is real and satisfies \(|\lambda|\leq1\).
	Moreover, \(k\geq1\) and \(b_0>\frac12\) imply \(b_k>1\).
	Since \(\varphi\) is positive on \((-b_k,b_k)\), we have
	\(\varphi(1)=\varphi(-1)>0\).
	
	Suppose that \(|\lambda|=1\), and let
	\(v=(v_j)_{j\in I_n}\neq0\) satisfy \(A_nv=\lambda v\).
	Set
	\[
	M:=\max_{j\in I_n}|v_j|>0,
	\]
	and choose \(i\in I_n\) such that \(|v_i|=M\). Then
	\[
	M
	=|\lambda v_i|=|(A_nv)_i|
	=\left|\sum_{j\in I_n}\varphi(i-j)v_j\right|
	\leq\sum_{j\in I_n}\varphi(i-j)|v_j|
	\leq M\sum_{j\in I_n}\varphi(i-j)
	\leq M.
	\]
	Thus all inequalities are equalities. In particular,
	\[
	\sum_{j\in I_n}\varphi(i-j)\bigl(M-|v_j|\bigr)=0.
	\]
	Since every summand is nonnegative, it follows that
	\[
	\varphi(i-j)>0
	\quad\Longrightarrow\quad
	|v_j|=M.
	\]
	
	If \(i<n\), then
	\(\varphi(i-(i+1))=\varphi(-1)>0\), so \(|v_{i+1}|=M\).
	Applying the same argument to the row \(i+1\) and proceeding
	successively through the remaining indices, we obtain
	\[
	|v_n|=M.
	\]
	However, by the partition-of-unity property,
	\[
	\sum_{j\in I_n}\varphi(n-j)
	\leq
	\sum_{j\in\mathbb Z}\varphi(n-j)-\varphi(-1)
	=1-\varphi(1)<1.
	\]
	Consequently,
	\[
	M
	=|\lambda v_n|=|(A_nv)_n|
	=\left|\sum_{j\in I_n}\varphi(n-j)v_j\right|
	\leq M\sum_{j\in I_n}\varphi(n-j)
	<M,
	\]
	a contradiction. Therefore every eigenvalue of \(A_n\)
	belongs to \((-1,1)\).
	
	Finally, since \(A_n\) is real and symmetric, it admits an
	orthogonal diagonalization
	\[
	A_n=Q_nD_nQ_n^{\mathsf T}.
	\]
	All diagonal entries of \(D_n\) have modulus strictly less
	than one, so \(D_n^\ell\to0\) as \(\ell\to\infty\). Hence
	\[
	A_n^\ell
	=Q_nD_n^\ell Q_n^{\mathsf T}
	\to 0,\,\, (n\to\infty).
	\]
\end{proof}

\begin{theorem}\label{thm:nn-iterates-zero}
	For every \(k,n\in\mathbb N\) and every \(f\in C[-1,1]\),
	\[
	\lim_{\ell\to\infty}
	\|\widetilde L_{n,k}^{\,\ell}f\|_\infty=0.
	\]
\end{theorem}

\begin{proof}
	With the notation above, Proposition
	\ref{prop:nn-matrix-spectrum} gives
	\(A_n^{\ell-1}\to0\) as \(\ell\to\infty\).
	The entries of \(U_n(x)\) are bounded uniformly for
	\(x\in[-1,1]\). Therefore
	\eqref{eq:nn-iterate-matrix} implies
	\[
	\sup_{x\in[-1,1]}
	|(\widetilde L_n^{\,\ell}f)(x)|
	\longrightarrow 0,
	\]
	as claimed.

\end{proof}

\begin{exam}\label{ex:nn-tridiagonal}
	Suppose that
	\[
	0<\varphi(1)<\varphi(0)<1,
	\qquad \varphi(2)=0.
	\]
	Then \(A_n\) is a tridiagonal matrix whose eigenvalues are
	\[
	\lambda_{n,r}
	=
	\varphi(0)-2\varphi(1)
	\cos\!\left(\frac{r\pi}{2n+2}\right),
	\qquad r=1,\ldots,2n+1;
	\]
	see \cite[Chapter~11, Exercise~38]{Bellman}.
	
	In this case, the partition-of-unity property gives
	\[
	\varphi(0)+2\varphi(1)=1,
	\]
	and it is readily verified that all eigenvalues belong
	to \((-1,1)\).
	
	In particular, if \(\varphi\) is the centered cardinal
	B-spline \(B_3\), then
	\[
	\varphi(0)=\frac34,
	\qquad
	\varphi(1)=\frac18.
	\]
\end{exam}

\section{Normalized neural network operators and their iterates}
\label{sec:normalized-nn-iterates}

Fix \(k\in\mathbb N\). We retain the notation of
Section~\ref{sec:nn-iterates} and write
\[
\varphi:=\varphi_k,\qquad
\sigma:=\sigma_k,\qquad
b:=b_k,\qquad
a:=a_k.
\]
In particular,
\[
\operatorname{supp}\varphi=[-b,b],
\qquad b=a+\frac12,
\]
and \(\varphi\) is positive on \((-b,b)\).

The operators \(\widetilde L_n\) do not, in general,
preserve constants on the whole interval \([-1,1]\).
To obtain operators with this property, we normalize
their weights. Set
\[
d_n(x):=(\widetilde L_n e_0)(x)
=\sum_{j\in I_n}\varphi(nx-j),
\qquad x\in[-1,1].
\]
By Proposition~\ref{prop:nn-constant},
\begin{equation}\label{eq:nn-normalizing-function}
	d_n(x)
	=
	\sigma\!\left(nx+n+\frac12\right)
	-
	\sigma\!\left(nx-n-\frac12\right).
\end{equation}

For every \(x\in[-1,1]\), there exists \(j_x\in I_n\)
such that \(|nx-j_x|\leq\frac12\). Since \(\varphi\) is
even and nonincreasing on \([0,\infty)\), we obtain
\[
d_n(x)\geq\varphi(nx-j_x)
\geq\varphi\!\left(\frac12\right)>0.
\]
On the other hand, the partition-of-unity property gives
\(d_n(x)\leq1\). Hence
\begin{equation}\label{eq:nn-normalizing-bounds}
	0<\varphi\!\left(\frac12\right)\leq d_n(x)\leq1,
	\qquad x\in[-1,1].
\end{equation}

We can therefore define the normalized operator
\[
L_n:C[-1,1]\longrightarrow C[-1,1]
\]
by
\begin{equation}\label{eq:normalized-nn-operator}
	(L_n f)(x):=
	\frac{(\widetilde L_n f)(x)}{d_n(x)}
	=
	\sum_{j\in I_n}\Phi_{n,j}(x)
	f\!\left(\frac{j}{n}\right),
\end{equation}
where
\begin{equation}\label{eq:normalized-nn-weights}
	\Phi_{n,j}(x):=\frac{\varphi(nx-j)}{d_n(x)},
	\qquad j\in I_n,\quad x\in[-1,1].
\end{equation}
These weights are continuous and satisfy
\[
\Phi_{n,j}(x)\geq0,
\qquad
\sum_{j\in I_n}\Phi_{n,j}(x)=1.
\]
Thus \(L_n\) is a positive linear operator and
\(L_n e_0=e_0\).

To describe its iterates, we first evaluate the weights
at the sampling nodes. For \(i\in I_n\), set
\begin{equation}\label{eq:normalized-nn-discrete-sums}
	d_{n,i}:=d_n\!\left(\frac{i}{n}\right)
	=\sum_{j\in I_n}\varphi(i-j).
\end{equation}
Then
\[
\Phi_{n,j}\!\left(\frac{i}{n}\right)
=\frac{\varphi(i-j)}{d_{n,i}}.
\]
Consequently,
\begin{equation}\label{eq:normalized-nn-weight-image}
	(L_n\Phi_{n,j})(x)
	=
	\sum_{i\in I_n}
	\frac{\varphi(i-j)}{d_{n,i}}\Phi_{n,i}(x).
\end{equation}

Introduce the matrix
\begin{equation}\label{eq:normalized-nn-matrix}
	M_n:=
	\left(\frac{\varphi(i-j)}{d_{n,i}}\right)_{i,j\in I_n}.
\end{equation}
Since its entries are nonnegative and each row sums
to one, \(M_n\) is stochastic. Moreover, the evenness
of \(\varphi\) implies
\[
d_{n,-i}=d_{n,i},
\qquad
(M_n)_{-i,-j}=(M_n)_{i,j},
\qquad i,j\in I_n.
\]
Hence \(M_n\) is centrosymmetric.

Using the row vector
\begin{equation}\label{eq.v1}
\boldsymbol{\Phi}_n(x)
:=
\bigl(\Phi_{n,-n}(x),\ldots,\Phi_{n,n}(x)\bigr)
\end{equation}
and the column vector \(F_n(f)\) defined in
Section~\ref{sec:nn-iterates}, we can write
\eqref{eq:normalized-nn-operator} as
\[
(L_n f)(x)=\boldsymbol{\Phi}_n(x)F_n(f).
\]
Using the sampling nodes, one gets
\[
\left\{
\begin{aligned}
	(L_n f)\left(\frac{-n}{n}\right)
	&= \Phi_{n,-n}\left(\frac{-n}{n}\right)
	f\left(\frac{-n}{n}\right)
	+ \cdots
	+ \Phi_{n,n}\left(\frac{-n}{n}\right)
	f\left(\frac{n}{n}\right), \\[4pt]
	&\quad \cdots \\[4pt]
	(L_n f)\left(\frac{n}{n}\right)
	&= \Phi_{n,-n}\left(\frac{n}{n}\right)
	f\left(\frac{-n}{n}\right)
	+ \cdots
	+ \Phi_{n,n}\left(\frac{n}{n}\right)
	f\left(\frac{n}{n}\right).
\end{aligned}
\right.
\]
These relations can be written as
\[
F_n(L_n f)
=
\begin{pmatrix}
	\Phi_{n,-n}\left(\dfrac{-n}{n}\right)
	& \cdots &
	\Phi_{n,n}\left(\dfrac{-n}{n}\right) \\[6pt]
	\vdots & & \vdots \\[4pt]
	\Phi_{n,-n}\left(\dfrac{n}{n}\right)
	& \cdots &
	\Phi_{n,n}\left(\dfrac{n}{n}\right)
\end{pmatrix}
\begin{pmatrix}
	f\left(\dfrac{-n}{n}\right) \\[6pt]
	\vdots \\[4pt]
	f\left(\dfrac{n}{n}\right)
\end{pmatrix}.
\]
Therefore,
\begin{equation*}
	F_n(L_n f)=M_n F_n(f).
	\end{equation*}
Iteration therefore yields
\begin{equation}\label{eq:normalized-nn-iterate-matrix}
	(L_n^\ell f)(x)
	=
	\boldsymbol{\Phi}_n(x)M_n^{\ell-1}F_n(f),
	\qquad
	\ell\in\mathbb N,\quad x\in[-1,1].
\end{equation}

We next identify the nodes at which normalization
leaves the original weights unchanged. Set
\[
\rho:=\lceil b\rceil-1
=
\begin{cases}
	\lfloor b\rfloor, & b\notin\mathbb N,\\
	b-1, & b\in\mathbb N.
\end{cases}
\]
Thus \(\rho\) is the largest nonnegative integer
strictly less than \(b\).

\begin{lemma}\label{lem:normalized-nn-row-sums}
	Let \(n\geq b\). Then
	\(\rho\in\{0,1,\ldots,n-1\}\), and
	\[
	d_{n,i}
	\begin{cases}
		=1, & -n+\rho\leq i\leq n-\rho,\\
		\in(0,1), & i\in I_n,\quad |i|>n-\rho.
	\end{cases}
	\]
\end{lemma}

\begin{proof}
	Since \(0\leq\rho<b\leq n\), we have
	\(\rho\in\{0,1,\ldots,n-1\}\).
	
	The support and positivity properties of \(\varphi\)
	give, for every integer \(r\),
	\[
	\varphi(r)>0
	\quad\Longleftrightarrow\quad
	|r|\leq\rho.
	\]
	Therefore the nonzero terms in
	\[
	\sum_{j\in\mathbb Z}\varphi(i-j)=1
	\]
	are precisely those with
	\(j\in\{i-\rho,\ldots,i+\rho\}\).
	
	If \(-n+\rho\leq i\leq n-\rho\), all these indices
	belong to \(I_n\). Hence no nonzero term is omitted
	from the sum defining \(d_{n,i}\), and \(d_{n,i}=1\).
	
	If \(i<-n+\rho\), the index \(j=i-\rho\) lies outside
	\(I_n\), and its contribution \(\varphi(\rho)>0\)
	is omitted. Similarly, if \(i>n-\rho\), the index
	\(j=i+\rho\) lies outside \(I_n\), and its contribution
	\(\varphi(-\rho)>0\) is omitted. Thus \(d_{n,i}<1\)
	in both cases. Finally, the term \(j=i\) gives
	\[
	d_{n,i}\geq\varphi(0)>0,
	\]
	which completes the proof.
\end{proof}

We now determine the limit of the powers of \(M_n\).
Throughout the following results, assume that
\(n\geq b\) and \(\rho>0\), and set
\[
d_n:=\sum_{i\in I_n}d_{n,i}.
\]
Here \(d_n\) denotes a scalar, whereas \(d_n(x)\) denotes
the normalizing function introduced above.

\begin{theorem}\label{thm:normalized-nn-matrix-limit}
	The powers of \(M_n\) converge to the matrix
	\begin{equation}\label{eq:normalized-nn-matrix-limit}
		B_n:=\lim_{\ell\to\infty}M_n^\ell
		=
		\frac{1}{d_n}
		\begin{pmatrix}
			d_{n,-n} & d_{n,-n+1} & \cdots & d_{n,n}\\
			d_{n,-n} & d_{n,-n+1} & \cdots & d_{n,n}\\
			\vdots & \vdots & & \vdots\\
			d_{n,-n} & d_{n,-n+1} & \cdots & d_{n,n}
		\end{pmatrix}.
	\end{equation}
\end{theorem}

\begin{proof}
	Since \(\rho>0\), we have \(1<b\), and therefore
	\(\varphi(1)>0\). Thus the diagonal entries of \(M_n\)
	and the entries immediately above and below the diagonal
	are strictly positive.
	
	In particular, \(M_n\) is primitive. Indeed, any two
	indices in \(I_n\) can be connected by at most \(2n\)
	steps between consecutive indices. The positive diagonal
	entries allow this path to be extended to exactly \(2n\)
	steps. Consequently, every entry of \(M_n^{2n}\) is
	strictly positive.
	
	Next, for every \(j\in I_n\), the evenness of \(\varphi\)
	gives
	\[
	\begin{aligned}
		\sum_{i\in I_n}d_{n,i}(M_n)_{i,j}
		&=
		\sum_{i\in I_n}\varphi(i-j)\\
		&=
		\sum_{i\in I_n}\varphi(j-i)
		=d_{n,j}.
	\end{aligned}
	\]
	Hence the probability row vector
	\begin{equation}\label{eq.v2}
	\pi_n:=
	\frac{1}{d_n}
	\bigl(d_{n,-n},d_{n,-n+1},\ldots,d_{n,n}\bigr)
	\end{equation}
	satisfies \(\pi_nM_n=\pi_n\).
	
	The convergence theorem for primitive stochastic matrices
	now yields
	\begin{equation*}
	M_n^\ell\longrightarrow\boldsymbol{1}\pi_n,
	\end{equation*}
	where \(\boldsymbol{1}\) is the column vector whose entries
	are all equal to one. This is precisely
	\eqref{eq:normalized-nn-matrix-limit}.
\end{proof}

\begin{theorem}\label{thm:normalized-nn-iterate-limit}
	For every \(f\in C[-1,1]\),
	\[
	\lim_{\ell\to\infty}(L_n^\ell f)(x)
	=
	\frac{1}{d_n}
	\sum_{i\in I_n}d_{n,i}f\!\left(\frac{i}{n}\right),
	\]
	uniformly for \(x\in[-1,1]\).
\end{theorem}

\begin{proof}
	By \eqref{eq:normalized-nn-iterate-matrix} and
	Theorem~\ref{thm:normalized-nn-matrix-limit},
	\[
	(L_n^\ell f)(x)
	=
	\boldsymbol{\Phi}_n(x)M_n^{\ell-1}F_n(f)
	\]
	Since
	\[
	\boldsymbol{\Phi}_n(x)\boldsymbol{1}
	=
	\sum_{j\in I_n}\Phi_{n,j}(x)=1,
	\]
	we have
	\[
	\boldsymbol{\Phi}_n(x)B_nF_n(f)
	=
	\pi_nF_n(f)
	=
	\frac{1}{d_n}
	\sum_{i\in I_n}d_{n,i}f\!\left(\frac{i}{n}\right).
	\]
	Moreover, the nonnegativity of the weights and their
	sum being one, together with (\ref{eq:normalized-nn-matrix-limit}),  imply
	\[
	\sup_{x\in[-1,1]}
	\left|
	\boldsymbol{\Phi}_n(x)
	\bigl(M_n^{\ell-1}-B_n\bigr)F_n(f)
	\right|
	\leq
	\left\|
	\bigl(M_n^{\ell-1}-B_n\bigr)F_n(f)
	\right\|_\infty
	\longrightarrow0.
	\]
	Thus the convergence is uniform.
\end{proof}

\section{The invariant measure of \(L_n\)}
\label{sec:nn-invariant-measure}

A probability Radon measure \(\mu\) on \([-1,1]\) is
invariant under \(L_n\) if
\[
\mu(L_n f)=\mu(f),
\qquad f\in C[-1,1],
\]
where
\[
\mu(f):=\int_{[-1,1]}f\,d\mu.
\]
Under the assumptions \(n\geq b\) and \(\rho>0\),
the preceding results give an explicit description
of the unique invariant probability measure.

\begin{theorem}\label{thm:nn-invariant-measure}
	The operator \(L_n\) has a unique invariant probability
	Radon measure, namely
	\[
	\mu_n
	=
	\frac{1}{d_n}
	\sum_{i\in I_n}d_{n,i}\delta_{i/n},
	\]
	where \(\delta_{i/n}\) denotes the Dirac measure at \(i/n\).
	Equivalently,
	\begin{equation}\label{eq:nn-invariant-measure}
		\mu_n(f)
		=
		\frac{1}{d_n}
		\sum_{i\in I_n}d_{n,i}f\!\left(\frac{i}{n}\right),
		\qquad f\in C[-1,1].
	\end{equation}
\end{theorem}

\begin{proof}
	The coefficients in the definition of \(\mu_n\) are
	positive and sum to one, so \(\mu_n\) is a probability
	Radon measure. Furthermore,
	\[
	\begin{aligned}
		\mu_n(L_n f)
		&=
		\frac{1}{d_n}
		\sum_{i\in I_n}d_{n,i}
		\sum_{j\in I_n}
		\frac{\varphi(i-j)}{d_{n,i}}
		f\!\left(\frac{j}{n}\right)\\
		&=
		\frac{1}{d_n}
		\sum_{j\in I_n}
		f\!\left(\frac{j}{n}\right)
		\sum_{i\in I_n}\varphi(i-j)\\
		&=
		\frac{1}{d_n}
		\sum_{j\in I_n}d_{n,j}f\!\left(\frac{j}{n}\right)
		=\mu_n(f).
	\end{aligned}
	\]
	Thus \(\mu_n\) is invariant.
	
	Let \(\mu\) be any invariant probability Radon measure.
	Iteration gives
	\[
	\mu(f)=\mu(L_n^\ell f),
	\qquad \ell\in\mathbb N.
	\]
	By Theorem~\ref{thm:normalized-nn-iterate-limit},
	\(L_n^\ell f\) converges uniformly to the constant
	function \(\mu_n(f)e_0\). Since \(\mu(e_0)=1\), we obtain
	\[
	\mu(f)
	=
	\lim_{\ell\to\infty}\mu(L_n^\ell f)
	=
	\mu\bigl(\mu_n(f)e_0\bigr)
	=
	\mu_n(f).
	\]
	Therefore \(\mu=\mu_n\), proving uniqueness.
\end{proof}

\section{The associated Markov chain}
\label{sec:nn-markov-chain}

We retain the notation of
Section~\ref{sec:normalized-nn-iterates} and assume that
\(n\geq b\) and \(\rho>0\).

For each fixed \(x\in[-1,1]\), consider the Markov chain
\((X_r)_{r\geq0}\) with state space
\[
S_n:=\left\{\frac{j}{n}:j\in I_n\right\},
\]
initial distribution
\[
\mathbb P_x\!\left(X_0=\frac{j}{n}\right)
=\Phi_{n,j}(x),
\qquad j\in I_n,
\]
and transition probabilities
\[
\mathbb P_x\!\left(
X_{r+1}=\frac{j}{n}\,\middle|\,X_r=\frac{i}{n}
\right)
=(M_n)_{i,j}
=\frac{\varphi(i-j)}{d_{n,i}},
\qquad i,j\in I_n.
\]
Here \(\boldsymbol{\Phi}_n(x)\) and \(M_n\) are defined
in \eqref{eq.v1} and \eqref{eq:normalized-nn-matrix},
respectively. We use standard terminology for finite
Markov chains; see \cite{LPW}.

\begin{theorem}\label{thm:nn-markov-chain}
	The associated Markov chain is irreducible,
	aperiodic, and reversible. Its unique stationary
	distribution is
	\[
	\pi_n
	=\frac{1}{d_n}
	\bigl(d_{n,-n},d_{n,-n+1},\ldots,d_{n,n}\bigr),
	\qquad
	d_n=\sum_{i\in I_n}d_{n,i}.
	\]
\end{theorem}

\begin{proof}
	Since \(\rho>0\), we have \(\varphi(1)>0\).
	Consequently,
	\[
	(M_n)_{i,i+1}>0,
	\qquad
	(M_n)_{i+1,i}>0,
	\qquad -n\leq i<n.
	\]
	Thus any two states can be connected by a sequence
	of transitions between consecutive states, proving
	irreducibility.
	
	Moreover,
	\[
	(M_n)_{i,i}
	=\frac{\varphi(0)}{d_{n,i}}>0,
	\qquad i\in I_n.
	\]
	Hence every state has period one, so the chain is
	aperiodic.
	
	Writing \(\pi_{n,i}:=d_{n,i}/d_n\), the evenness
	of \(\varphi\) yields
	\[
	\pi_{n,i}(M_n)_{i,j}
	=\frac{\varphi(i-j)}{d_n}
	=\frac{\varphi(j-i)}{d_n}
	=\pi_{n,j}(M_n)_{j,i},
	\qquad i,j\in I_n.
	\]
	These are the detailed balance equations.
	Therefore the chain is reversible with respect
	to \(\pi_n\). Summing over \(i\), we obtain
	\[
	\sum_{i\in I_n}\pi_{n,i}(M_n)_{i,j}
	=\pi_{n,j}\sum_{i\in I_n}(M_n)_{j,i}
	=\pi_{n,j},
	\]
	so \(\pi_n\) is stationary. Its uniqueness follows
	from irreducibility.
\end{proof}

\begin{remark}\label{rem:nn-stationary-uniform}
	By Lemma~\ref{lem:normalized-nn-row-sums} and the
	symmetry \(d_{n,-i}=d_{n,i}\), the stationary
	distribution has the form
	\[
	\pi_n=\frac{1}{d_n}
	\bigl(
	d_{n,n},d_{n,n-1},\ldots,d_{n,n-\rho+1},
	\underbrace{1,\ldots,1}_{2n+1-2\rho},
	d_{n,n-\rho+1},\ldots,d_{n,n-1},d_{n,n}
	\bigr).
	\]
	Moreover,
	\[
	0<d_{n,n}<d_{n,n-1}<\cdots
	<d_{n,n-\rho+1}<1.
	\]
	Indeed, for \(s=0,\ldots,\rho\), the support
	property of \(\varphi\) and the assumption \(n\geq b\)
	give
	\[
	d_{n,n-s}
	=\sum_{r=-s}^{\rho}\varphi(r).
	\]
	Consequently,
	\[
	d_{n,n-s-1}-d_{n,n-s}
	=\varphi(s+1)>0,
	\qquad s=0,\ldots,\rho-1,
	\]
	and \(d_{n,n-\rho}=1\).
	
	In particular,
	\[
	d_n
	=2\sum_{s=0}^{\rho-1}d_{n,n-s}
	+2n+1-2\rho,
	\]
	so
	\begin{equation}\label{eq.r}
	2n+1-2\rho<d_n<2n+1,
	\end{equation}
	or, equivalently,
	\[
	\frac{1}{2n+1}
	<\frac{1}{d_n}
	<\frac{1}{2n+1-2\rho}.
	\]
	
	For a fixed kernel \(\varphi\), the distribution
	\(\pi_n\) approaches the uniform distribution
	\[
	u_n:=\frac{1}{2n+1}(1,\ldots,1)
	\]
	as \(n\to\infty\). More precisely, consider the total variation distance (see \cite[Prop.4.2]{LPW}),
	\[
	\|\pi_n-u_n\|_{\mathrm{TV}}
	:=\frac12\sum_{i\in I_n}
	\left|\pi_{n,i}-\frac{1}{2n+1}\right|,
	\]
	and define the probability vector
	\[
	v_n:=
	\left(
	\frac{1-d_{n,i}}{2n+1-d_n}
	\right)_{i\in I_n}.
	\]
	Then
	\[
	u_n
	=\frac{d_n}{2n+1}\pi_n
	+\frac{2n+1-d_n}{2n+1}v_n.
	\]
	Therefore,
	\[
	\begin{aligned}
		\lVert \pi_n-u_n\rVert_{\mathrm{TV}}
		&=
		\left\lVert
		\pi_n-\frac{d_n}{2n+1}\pi_n
		-\frac{2n+1-d_n}{2n+1}v_n
		\right\rVert_{\mathrm{TV}} \\[4pt]
		&=
		\left\lVert
		\frac{2n+1-d_n}{2n+1}\pi_n
		-\frac{2n+1-d_n}{2n+1}v_n
		\right\rVert_{\mathrm{TV}} \\[4pt]
		&=
		\frac{2n+1-d_n}{2n+1}
		\lVert \pi_n-v_n\rVert_{\mathrm{TV}} \\[4pt]
		&\le
		\frac{2n+1-d_n}{2n+1}
		\left(
		{\lVert \pi_n\rVert_{\mathrm{TV}}}
		+
		{\lVert v_n\rVert_{\mathrm{TV}}}
		\right) \\[4pt]
		&= \frac{2n+1-d_n}{2n+1}
		<
		\frac{2\rho}{2n+1}.
	\end{aligned}
	\]
	So, we have the estimate $	\lVert \pi_n-u_n\rVert_{\mathrm{TV}}\leq \dfrac{2\rho}{2n+1}\to 0$, $n\to\infty$. 
\end{remark}

\begin{exam}\label{ex:nn-birth-death}
	Suppose that \(1<b\leq2\). Then
	\[
	\varphi(1)>0,
	\qquad
	\varphi(r)=0
	\quad\text{for every integer } |r|\geq2,
	\]
	and the partition-of-unity property gives
	\[
	\varphi(0)+2\varphi(1)=1.
	\]
	
	Relabeling the state \(i/n\) by \(s=i+n\),
	we obtain a birth-and-death chain on
	\(\{0,1,\ldots,2n\}\); see \cite{LPW}.
	Let \(p_s\), \(q_s\), and \(r_s\) denote,
	respectively, the probabilities of moving from
	\(s\) to \(s+1\), moving from \(s\) to \(s-1\),
	and remaining at \(s\).
	
	At the boundary states, these probabilities are
	\[
	p_0=q_{2n}
	=\frac{\varphi(1)}{\varphi(0)+\varphi(1)},
	\qquad
	q_0=p_{2n}=0,
	\]
	and
	\[
	r_0=r_{2n}
	=\frac{\varphi(0)}{\varphi(0)+\varphi(1)}.
	\]
	At the interior states,
	\[
	p_s=q_s=\varphi(1),
	\qquad
	r_s=\varphi(0),
	\qquad s=1,\ldots,2n-1.
	\]
	
	Thus the transition matrix is tridiagonal, with
	entries
	\[
	(M_n)_{s,t}
	=
	\begin{cases}
		\dfrac{\varphi(0)}{\varphi(0)+\varphi(1)},
		& s=t\in\{0,2n\},\\[2mm]
		\varphi(0),
		& s=t,\quad 1\leq s\leq2n-1,\\[1mm]
		\dfrac{\varphi(1)}{\varphi(0)+\varphi(1)},
		& (s,t)\in\{(0,1),(2n,2n-1)\},\\[2mm]
		\varphi(1),
		& |s-t|=1,\quad 1\leq s\leq2n-1,\\[1mm]
		0, & \text{otherwise}.
	\end{cases}
	\]
	Here the indices \(s,t\in\{0,\ldots,2n\}\)
	refer to the relabeled states. This is precisely
	the matrix in \eqref{eq:normalized-nn-matrix},
	since
	\[
	d_{n,-n}=d_{n,n}=\varphi(0)+\varphi(1),
	\qquad
	d_{n,i}=1
	\quad\text{for }-n<i<n.
	\]
\end{exam}

\begin{cor}\label{cor:nn-markov-ergodic}
	Under the assumptions of
	Theorem~\ref{thm:nn-markov-chain}, for every
	\(x\in[-1,1]\) and every \(f\in C[-1,1]\),
	\[
	\lim_{N\to\infty}
	\frac{1}{N}\sum_{r=0}^{N-1}f(X_r)
	=
	\mu_n(f)
	=
	\frac{1}{d_n}
	\sum_{i\in I_n}d_{n,i}f\!\left(\frac{i}{n}\right),
	\qquad \mathbb P_x\text{-almost surely}.
	\]
	In particular, for every \(i\in I_n\),
	\[
	\lim_{N\to\infty}
	\frac{1}{N}\sum_{r=0}^{N-1}
	\mathbf{1}_{\{X_r=i/n\}}
	=
	\frac{d_{n,i}}{d_n},
	\qquad \mathbb P_x\text{-almost surely}.
	\]
\end{cor}

\begin{proof}
	By Theorem~\ref{thm:nn-markov-chain}, the chain is
	finite and irreducible, with stationary distribution
	\(\pi_{n,i}=d_{n,i}/d_n\). The ergodic theorem,
	applied to the restriction of \(f\) to \(S_n\),
	therefore gives
	\[
	\frac{1}{N}\sum_{r=0}^{N-1}f(X_r)
	\longrightarrow
	\sum_{i\in I_n}\pi_{n,i}f\!\left(\frac{i}{n}\right)
	=\mu_n(f)
	\]
	almost surely, for any initial distribution.
	The second assertion follows by applying the same
	theorem to the indicator of the state \(i/n\).
\end{proof}

\begin{remark}\label{rem:nn-probabilistic-iterates}
	The distribution of \(X_r\) is
	\(\boldsymbol{\Phi}_n(x)M_n^r\). Consequently,
	\eqref{eq:normalized-nn-iterate-matrix} yields
	\[
	\mathbb E_x[f(X_r)]
	=
	\boldsymbol{\Phi}_n(x)M_n^rF_n(f)
	=
	(L_n^{r+1}f)(x),
	\qquad r\geq0.
	\]
	Thus Theorem~\ref{thm:normalized-nn-iterate-limit}
	and Corollary~\ref{cor:nn-markov-ergodic} provide
	two probabilistic descriptions of the same
	quantity \(\mu_n(f)\):
	\[
	\lim_{r\to\infty}\mathbb E_x[f(X_r)]
	=\mu_n(f),
	\qquad
	\lim_{N\to\infty}
	\frac{1}{N}\sum_{r=0}^{N-1}f(X_r)
	=\mu_n(f)
	\quad \mathbb P_x\text{-almost surely}.
	\]
\end{remark}

%\section*{Funding}
%............

\section*{Conflict of interest/Competing interests}

{\small The authors declare that they have no conflict of interest and competing interest.}

$  $

\end{document}